\documentclass[twocolumn]{article}

\usepackage{hyperref}
\usepackage{tikz}
\usetikzlibrary{chains,shapes.multipart}
\usetikzlibrary{shapes}
\usetikzlibrary{automata,positioning}
\usetikzlibrary{calc}
\usepackage{tikzscale}
\usepackage{xcolor}
\usepackage{makecell}
\usepackage{tabularx}
\usepackage{tkz-graph}
\usepackage{float}
\usepackage{amsfonts}
\usetikzlibrary{shapes.geometric}
\usepackage{amsmath}
\usepackage{amssymb}
\usepackage{braket}
\usepackage{graphicx}
\usepackage{colortbl}
\usepackage{comment}
\usepackage{multirow}
\usepackage{url}
\usepackage{soul}
\usepackage{geometry}
 \usepackage{hhline}

\AtBeginDocument{%
  \providecommand\BibTeX{{%
    \normalfont B\kern-0.5em{\scshape i\kern-0.25em b}\kern-0.8em\TeX}}}

\newcommand{\mainfindingbox}[1]{%
            \begin{tikzpicture}%
                \node[rectangle, draw=black, style=thick, fill=gray!20, rounded corners=2pt, inner xsep=5pt, inner ysep=6pt, outer ysep=10pt]{
                \begin{minipage}{0.96\linewidth}#1\end{minipage}};%
            \end{tikzpicture}
}

\newcommand{\observationbox}[1]{%
            \begin{tikzpicture}%
                \node[rectangle, draw=black, rounded corners=2pt, inner xsep=5pt, inner ysep=6pt, outer ysep=10pt]{
                \begin{minipage}{0.96\linewidth}#1\end{minipage}};%
            \end{tikzpicture}%
}

\title{On Optimizing Hyperparameters for Quantum Neural Networks}
\author{Sabrina Herbst \and Vincenzo De Maio \and Ivona Brandic\thanks{Vienna University of Technology, Vienna, Austria, sabrina.herbst@tuwien.ac.at, \{vincenzo,ivona\}@ec.tuwien.ac.at}}
\date{}

\begin{document}

\maketitle

\begin{abstract} 
The increasing capabilities of Machine Learning (ML) models go hand in hand with an immense amount of data and computational power required for training. Therefore, training is usually outsourced into HPC facilities, where we have started to experience limits in scaling conventional HPC hardware, as theorized by Moore's law. Despite heavy parallelization and optimization efforts, current state-of-the-art ML models require weeks for training, which is associated with an enormous $CO_2$ footprint. Quantum Computing, and specifically Quantum Machine Learning (QML), can offer significant theoretical speed-ups and enhanced expressive power. However, training QML models requires tuning various hyperparameters, which is a nontrivial task and suboptimal choices can highly affect the trainability and performance of the models. In this study, we identify the most impactful hyperparameters and collect data about the performance of QML models. We compare different configurations and provide researchers with 
performance data and concrete suggestions for hyperparameter selection.

\end{abstract}

\section{Introduction}

With the exponential increase of data of recent years, Machine Learning (ML) has become a key technology for analytics and pattern recognition~\cite{gandomi_machine_2022}. Beyond that, ML models are becoming prevalent in everyday life, as, for example, Large Language Models (LLMs) assist in day-to-day tasks. However, these models require a lot of training time and are associated with a significant $CO_2$ footprint. Meta's Llama 2 LLMs require between 184,320 and 1,720,320 GPU hours for pretraining, which is associated with between 31.22 and 291.42 $tCO_2$ equivalents~\cite{touvron2023llama} (as a comparison, the EU27 per-capita fossil fuel emissions for 2021 were 6.25 $tCO_2$ equivalents~\cite{edgar_data}). These factors, together with the recently reached limit of the Von Neumann architecture, motivate the need to %
scale computational and storage resources to meet the increasing demand of modern ML applications. The most promising path is the integration of Non-Von Neumann architectures, such as Quantum Computing, into the HPC continuum~\cite{Schulz2021}.%

Quantum computers exploit principles of quantum mechanics, resulting in speed-ups over classical computers for certain computations, and pave the way to entirely novel solutions~\cite{gyongyosi_survey_2019}. Recent quantum computing advances, and the accessibility of quantum machines in the cloud, have opened the doors to quantum computing for researchers across various disciplines~\cite{de_luca_survey_2022}. Quantum Machine Learning (QML), i.e., extending the learning paradigm to quantum computers, %
has emerged as a promising area of research, which could offer significant advantages in terms of runtime, performance, and space efficiency~\cite{pastorello_concise_2023}.

However, several problems limit the adoption of QML. First, the small number of qubits that is available today limits the input data. Moreover, only Noisy Intermediate-Scale Quantum (NISQ) technology will be available in the near future~\cite{preskill_quantum_2018}, meaning current and near-term quantum hardware exhibits significant levels of noise, resulting in errors during computation~\cite{resch_benchmarking_2022}. Furthermore, encoding classical data into a quantum state can be computationally intensive and could cancel out potential runtime benefits~\cite{sharma_quantum_2021}. Finally, the model architecture can strongly affect its performance and determining the optimal choice poses an optimization problem with an intractable search space~\cite{schuld_is_2022}. To this date, there exist only few studies, with limited scope, investigating these hyperparameters.

\textbf{Our contribution.} We collect data about QML models by investigating the influence of different hyperparameters on their performance. We select four classical classification datasets, identify the main hyperparameters, and evaluate their impact on the runtime and predictive performance of the models. Further, we study the impact of hardware noise on the different hyperparameters. 

\textbf{We focus on Quantum Neural Networks (QNNs)}, due to their %
relevance for near-term hardware~\cite{cerezo_variational_2021}, and on tabular data as the number of available quantum bits (qubits) limits the input data. %
 We employ the IBM Qiskit package and the included simulators~\cite{Qiskit} for our evaluation. 
We choose Qiskit as a programming language, as it is a standard for publicly available state-of-the-art quantum machines, which can be used with quantum backends besides IBM, such as Amazon\footnote{\url{https://github.com/qiskit-community/qiskit-braket-provider}. Accessed: 04.01.2024} or AQT\footnote{\url{https://github.com/qiskit-community/qiskit-aqt-provider}. Accessed: 04.01.2024}. One caveat of using the package is that it allows only limited configuration options, reducing the possibility of adapting the models. Nonetheless, given the limited accessibility of quantum computers, this work will support scientists and practitioners in furthering research in QML with these very machines by providing data and guidelines for hyperparameter tuning. To the best of our knowledge, this work constitutes the biggest publicly available study for QNN architectures to this date.

Our results show that the \textbf{optimizer and initialization method constitute the most important hyperparameters for QNNs}. Entangling the feature map can be favorable, but requires efficient initialization to overcome issues during training. The exact entangling strategy has a negligible effect on the performance of the model in our setting. All code and results are available publicly\footnote{\url{https://github.com/sabrinaherbst/hyperparameters_qnn}}.

We structure the paper as follows: first, we provide preliminary concepts and related work in Section~\ref{sec:background}. %
In Section~\ref{benchmarking_methodology} we describe our methodology. The experimental setup and evaluation are discussed respectively in Section~\ref{sec:experimental} and Section~\ref{sec:evaluation}. Our experimental results are discussed in Section~\ref{sec:discussion}, including threats to validity of this work. Finally, we conclude our paper in Section~\ref{sec:conclusion}.

\section{Background}
\label{sec:background}
\subsection{Quantum Information Theory}
Classical computers work with bits, which are either in state 0 or 1. In quantum computing, the so-called \emph{computational basis} is composed of the states $\ket{0}$ and $\ket{1}$. A single-qubit quantum state $\ket{\phi}$ forms a linear combination of these two states, called a \emph{superposition}, represented as $\ket{\phi} = \alpha\ket{0} + \beta\ket{1}$, with $\alpha, \beta \in \mathbb{C}$ and $|\alpha|^2 + |\beta|^2 = 1$. $\ket{\phi}$ is in both states at the same time, but takes the value of $\ket{0}$ with probability $|\alpha|^2$ and $\ket{1}$ with probability $|\beta|^2$, once we measure the state. \emph{Quantum parallelism} harnesses this phenomenon, allowing quantum computers to concurrently represent multiple states, rather than needing to evaluate each state separately. 

Another peculiarity of quantum computing is \emph{quantum entanglement}, where multiple qubits are correlated, such that performing an action on one qubit will impact all qubits it is entangled with as well. Superposition and entanglement are usually exploited in quantum algorithms to achieve significant speed-ups.

\subsection{Quantum Machine Learning}
\begin{figure}[!th]
    \centering
    \tikzstyle{startstop} = [circle, rounded corners, minimum width=0.1cm, minimum height=0.1cm,text centered, draw=black, double, fill=black]
    \tikzstyle{process} = [rectangle, minimum width=1cm, minimum height=0.3cm, text centered, draw=black, fill=blue!30]
    \tikzstyle{decision} = [diamond, minimum width=0.05cm, minimum height=0.05cm, text centered, draw=black, fill=red!30]
    \tikzstyle{quantum} = [rectangle, minimum width=1cm, minimum height=0.3cm, text centered, draw=black, fill=yellow!30]
    \tikzstyle{classical} = [rectangle, minimum width=1cm, minimum height=0.3cm, text centered, draw=black, fill=red!30]
    \tikzstyle{arrow} = [thick,->,>=stealth]
    \begin{tikzpicture}
    \node (start) [process] {\makecell[c]{Data \\ $x$}};
    \node (encoding) [process, right=.2cm of start] {\makecell[c]{Encoding\\
    $V(x)$}};
    \node (initial) [process, above=.3cm of encoding] {\makecell[c]{Initial \\
    $\vec{\Theta}$}};
    \node (circuit) [process, below=.3cm of encoding] {\makecell[c]{Circuit $U$}};
    \node (quantum) [quantum, right=.3cm of encoding] {
    \makecell[c]{
        $V(x) \otimes$\\
        $U(\vec{\Theta})$
    }
    };
    \node(evaluate)[classical, right=.3cm of quantum] {\makecell[c]{Evaluate\\
    $C(\vec{\Theta})$}};
    \node (decision) [decision,right=.3cm of evaluate] {\makecell[c]{\footnotesize End?}};
    \node (c-theta) [classical, below=.4cm of evaluate] {\makecell[c]{Optimize\\$\vec{\Theta}$}};
    \node (new-theta) [process, left=.4cm of c-theta] {\makecell[c]{New\\$\vec{\Theta}$}};
    \node (stop) [startstop, above=.6cm of decision]{};%
    \draw[arrow] (start) -- (encoding);
    \draw[arrow] (initial) -- (quantum);
    \draw[arrow] (encoding) -- (quantum);
    \draw[arrow] (circuit) -- (quantum);
    \draw[arrow] (quantum) -- (evaluate);
    \draw[arrow] (evaluate) -- (decision);
    \draw[arrow] (decision) |- node[anchor=west]{No} (c-theta);
    \draw[arrow] (c-theta) -- (new-theta);
    \draw[arrow] (new-theta) -- (quantum);
    \draw[arrow] (decision) -- node[anchor=east]{Yes} (stop);

    \matrix [draw,below=.15cm] at (current bounding box.south) {
      \node [process,label=right:Input] {}; \\
      \node [classical,label=right:Classical Computer] {}; \\
      \node [quantum,label=right:Quantum Computer] {}; \\
    };
    \end{tikzpicture}
    \caption{Quantum Neural Networks}
    \label{fig:vqa}
\end{figure}
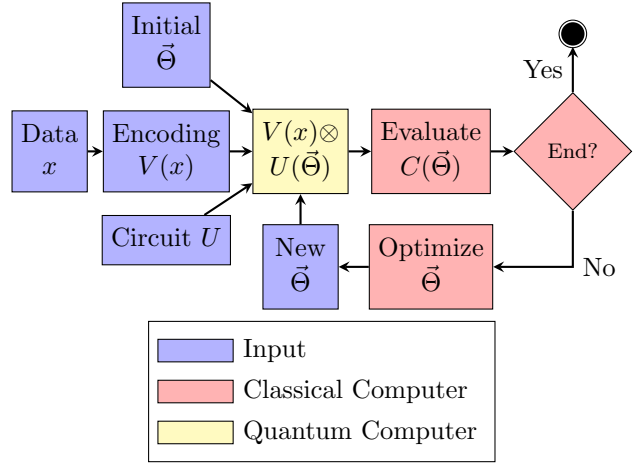

Working with classical data on quantum computers requires an effective strategy for encoding the information into a quantum state~\cite[p.25-28]{pastorello_concise_2023}. Exploiting the unique characteristics of a quantum computer increases the chance of obtaining an advantage, therefore, it is recommended to choose an encoding that is hard to obtain or simulate on a classical computer~\cite{goto_universal_2021}.

Similarly to classical ML, several classification algorithms exist, including Quantum Support Vector Machines~\cite{rebentrost_quantum_2014} or K-Means Clustering~\cite{lloyd2013quantum}, both with an exponential speed-up compared to the classical algorithms. However, Quantum Neural Networks (QNNs) have come up as one of the leading algorithms that could achieve a quantum advantage on NISQ technology, due to their simple architecture and training process~\cite{cerezo_variational_2021}. 

QNNs are depicted in Figure~\ref{fig:vqa}. Initially, classical data is encoded into a quantum state using a feature encoding method $V$. The inputs to the algorithm are a parametrized quantum circuit $U$ (ansatz), which is a sequence of trainable unitary transformations, and associated initial parameters $\vec{\Theta}$. The goal is to find parameters $\vec{\Theta}$ that minimize the cost function $C$, i.e., cross-entropy for classification. After executing the circuit, the parameters $\vec{\Theta}$ are optimized on a classical computer using an optimizer, such as ADAM~\cite{kingma2017adam} or COBYLA~\cite{powell_direct_1994}. The parameters are then transferred to the quantum computer, where the circuit is executed again. The process is repeated until a certain termination criterion is reached (i.e., maximum number of iterations, tolerance threshold). The different choices for feature map, ansatz and optimizer can highly impact the results.

Currently, one of the biggest challenges in QNNs is mitigating the \textbf{Barren Plateau (BP)} phenomenon. BPs cause the gradient of the cost function to vanish exponentially in problem size, rendering optimization of the parameters difficult~\cite{cerezo_variational_2021}. In particular, randomly initialized deep circuits are subject to these plateaus~\cite{bp_qnn}. This has been extended for shallow ones and linked to the structure of the cost function~\cite{cerezo_cost_bp}. Furthermore, BPs have been linked to expressivity~\cite{PRXQuantum.3.010313}, entanglement~\cite{entanglement_induced} and noise{~\cite{nibp}}. Ways to overcome them include different parameter initialization strategies{~\cite{PhysRevX.10.021067,Grant2019initialization,PRXQuantum.1.020319}}, or limiting the size of the accessible Hilbert space through, e.g., problem-specific ansatzes~\cite{verdon2019learning,a12020034}.

\subsection{Related Work}\label{related_work}
In~\cite{sierra-sosa_data_2023}, the influence of the type of data, classical transformations on data, and feature encoding techniques on QML models, focusing on amplitude encoding and the ZZFeatureMap, are investigated. The authors generate synthetic datasets using different transformations and apply various rotations to them. They observe that amplitude encoding is less affected by transformations than the ZZFeatureMap is. As rotating the dataset improves some models, the authors propose a strategy of rotating it to create a different representation. The approach aligns with the idea of~\cite{schuld_effect_2021}, suggesting that appropriately scaling the data may be beneficial in QML.

Mancilla and Pere employ Classical and Quantum ML in~\cite{mancilla_preprocessing_2022} and compare different feature reduction techniques. They find that Linear Discriminant Analysis (LDA) outperforms Principal Component Analysis (PCA) for QML models. Hancco-Quispe et al. report similar results in~\cite{hancco-quispe_quantum_2023}.

Joshi et al.~\cite{joshi_evaluating_2021} investigate local optimizers in QNNs using EfficientSU2 as the ansatz and ZFeatureMap as the feature map. They compare the performance using AQGD and COBYLA to classical machine learning models in analyzing sentiment data. The QNN with AQGD outperforms the other models.

Moreover, the authors in~\cite{suryotrisongko_hybrid_2022} compare the performances of a hybrid deep learning model and a QNN. They consider different feature map, ansatz and optimizer combinations for the QNN and find significant differences in performance.

Piatrenka and Rusek~\cite{piatrenka_quantum_2022} perform experiments using different configurations for the iris dataset. They experiment using ZFeatureMap, ZZFeatureMap and a PauliFeatureMap with custom Pauli gates. Regarding the ansatz, they only experiment using the RealAmplitudes one, but vary the repetitions. For the optimizer, they consider SPSA, COBYLA and SLSQP. They find that one repetition of the PauliFeatureMap and two repetitions of the RealAmplitudes ansatz work best. Furthermore, they report that COBYLA and SLSQP outperform SPSA. The authors run their experiments on simulators and a real quantum computer and find big performance differences between the executions.

Moreover, Katyayan and Joshi propose a model for classifying questions in~\cite{katyayan_supervised_2023}. They use a TwoLocal ansatz and experiment with all three of Qiskit's feature maps. They find that the PauliFeatureMap works best, that a small depth is advantageous, and that the features strongly influence the output.

Finally, Joshi et al. compare classical ML models to quantum ones on sentiment analysis in~\cite{joshi_comparing_2021}. They find that quantum models slightly outperform the classical models in their configuration. They employ a PauliFeatureMap using custom Pauli gates and COBYLA as the optimizer. The EfficientSU2 ansatz with 100 epochs outperforms the other models.

Other works, such as~\cite{autoqml_1, autoqml_2} define Automated Machine Learning (AutoML) for QML by proposing algorithms for automatically tuning the QML hyperparameters.

\section{Methodology}\label{benchmarking_methodology}
In the following, we describe our methodology, the chosen datasets and selected QNN hyperparameters. %

\subsection{Data}
We choose datasets with low dimensionality, due to the limited qubits on the IBM Quantum Open Access Plan\footnote{Note: The plan was changed after we conducted the experiments. IBM used to offer free seven-qubit-machines for the public, now larger machines are available but with a ten-minute time constraint per month.}. %

We consider (1) the KDD Cup 1999 dataset~\cite{lippmann_1999_2000}, a well-known public classification dataset for intrusion detection, (2) the Cover Type dataset, whose goal is to predict the cover type of forest squares based on cartographic attributes~\cite{misc_covertype_31}, (3) the Glass Identification dataset~\cite{misc_glass_identification_42}, used to identify types of glass for forensics at crime scenes, and (4) the Rice dataset, for binary classification between two types of rice~\cite{misc_rice_(cammeo_and_osmancik)_545, Cinar2019ClassificationOR}. All datasets are available on the UCI Machine Learning repository~\cite{UCIdatasets}. 

The datasets offer diverse characteristics. While the Rice dataset is only for binary classification, both Glass Identification and Cover Type have seven target categories and the KDD Cup dataset distinguishes between 22 different attacks. The Glass Identification dataset consists of only 214 samples, whereas the other datasets exceed our maximum training samples of 400. We apply a One-Hot Encoding for categorical features and compare PCA and LDA for reducing the dimensionality to seven features. We do not employ LDA for the Rice dataset, as it would lead to only one feature. As the encoding we choose employs one qubit per feature, entangling the qubits would not be possible anymore. %

\subsection{Optimizers}

We employ COBYLA~\cite{powell_direct_1994}, SPSA~\cite{spall_overview_1998} and Nelder-Mead~\cite{nelder_simplex_1965} as optimizers. Both COBYLA and SPSA are extensively used in the QML literature, as mentioned in Section~\ref{related_work}. We also select Nelder-Mead due to its popularity on different quantum optimization problems~\cite{Willsch2020,Moussa_2020,parameter_fixing}.

SPSA is a gradient-based method, which is known to work well in the presence of noise\footnote{\url{https://qiskit.org/documentation/stubs/qiskit.algorithms.optimizers.SPSA.html}. Accessed 17.06.2023}. It calculates the objective function using only two measurements, irrespective of the number of parameters involved in the optimization problem. Nelder-Mead is a heuristic gradient-free method which performs unconstrained optimization, and is based on the simplex algorithm. COBYLA is gradient-free as well, but uses trust regions instead and can be used for constrained optimization. By utilizing optimizers with diverse characteristics, we aim to identify which type of optimizer is likely to lead to high accuracy for QML models.  

\subsection{Ansatzes}
Ansatzes are characterized by the circuit structure, i.e., the sequence of operations and the employed entanglement. %
\subsubsection{Circuit Structure}

The ansatz defines the trainable parameters of the circuit. We select four popular ansatzes from the Qiskit package, which are widely used in QML~\cite{joshi_evaluating_2021, katyayan_supervised_2023, piatrenka_quantum_2022} and relevant for near-term hardware: PauliTwoDesign, RealAmplitudes, EfficientSU2 and TwoLocal. The ansatzes are built by repeatedly applying alternating rotation and entanglement blocks. The number of repetitions can be tuned and defines the circuit depth.

$R_{X|Y|Z}$ gates are used for rotations around the X, Y and Z axes of the Bloch sphere, which is a geometrical representation of a qubit, the angle defined by the learnable parameters $\vec{\theta}$. Furthermore, two qubits are entangled using entanglement gates, which are either controlled-X gates or controlled-Z gates. They perform an X/NOT operation ($\ket{0}\rightarrow\ket{1}$, $\ket{1}\rightarrow\ket{0}$, see Equation{~\ref{eq:x}}) or Z ($\ket{0}\rightarrow\ket{0}$, $\ket{1}\rightarrow -\ket{1}$, see Equation{~\ref{eq:z}}) transformation on one qubit (called the target), should the other one (control) be in state $\ket{1}$.

\noindent
\begin{minipage}{.45\linewidth}
\vspace{0.25cm}
\begin{equation}\label{eq:x}
    X = \begin{pmatrix} 0 & 1 \\ 1 & 0 \end{pmatrix}
\end{equation}
\vspace{0.25cm}
\end{minipage}
\begin{minipage}{.45\linewidth}
\vspace{0.25cm}
\begin{equation}\label{eq:z}
    Z = \begin{pmatrix} 1 & 0 \\ 0 & -1 \end{pmatrix}\\
\end{equation}
\vspace{0.25cm}
\end{minipage}

\begin{figure}
    \centering
    \includegraphics[width=\linewidth]{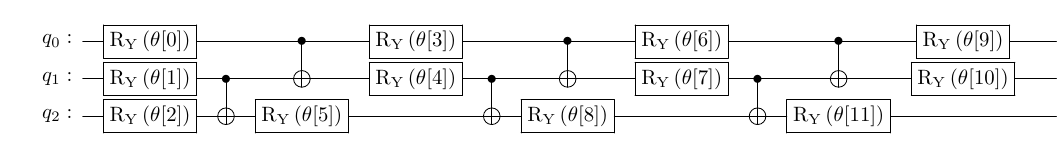}
    \caption{RealAmplitudes Ansatz}
    \label{fig:realamplitudes}
\end{figure}

For better illustration, we show the RealAmplitudes ansatz in Figure{~\ref{fig:realamplitudes}}. The $R_Y$ operations rotate the individual qubits around the y-axis of the Bloch sphere, with an angle defined by the trainable parameter $\theta$. The operations in between are the controlled-NOT gates (CNOT), which entangle the qubits. The target qubit is denoted as $\bigoplus$ and the control qubit as $\bullet$. The rotation and entanglement blocks are repeated three times in this case, making up an ansatz of depth three. The goal of training is to find the optimal angles $\vec{\theta}$.

We experiment with the entanglement hyperparameter, leaving all others to their default values. %
Additionally, for TwoLocal, as suggested in the Qiskit documentation, we utilize the $R_Y$ rotation and controlled-X entanglement block.

We experiment using Qiskit's default initialization (uniform in range $[0, 1]$), the beta distribution, as proposed in~\cite{beinit}, and the normal distribution with $\sigma^2=\frac{1}{L}$, where $L$ is the number of layers~\cite{NEURIPS2022_7611a3cb}.

\subsubsection{Qubit Entanglement}

Another design choice is how to entangle qubits. Possible choices are full (each qubit is entangled with all others), linear (qubit $q_i$ is entangled with $q_{i+1}$, with $i \in [0, N-2]$ where $N$ is the number of qubits), circular (linear, plus $q_{N-1}$ is entangled with $q_0$), or sca (shifted-circular-alternating). 

In sca entanglement, the target and control qubit of the entanglement gates are swapped in each block. Furthermore, the entanglement strategy is similar to circular. Qubits $q_{N-1}$ and $q_{0}$ are entangled in the first iteration before all others. In the next iteration, first $q_0$ and $q_1$ are entangled, then $q_{N-1}$ and $q_{0}$, and then the rest. The entanglement between $q_{N-1}$ and $q_0$ is "shifted" in every iteration. We exclude reverse-linear due to its similarity to the linear hyperparameter choice.

TwoLocal additionally supports pairwise entanglement (in even layers, qubit $q_i$ is entangled with $q_{i+1}$, in uneven ones with $q_{i-1}$).

\subsection{Encoding}
Working with classical data on quantum computers requires encoding the information into a quantum state. A very simple example of data encoding is the so-called \emph{basis encoding}. The value, for example the number $5$, is transformed into a binary string ($101$) and each bit is translated into the quantum counterpart, requiring three qubits: $5 \rightarrow 101 \rightarrow \ket{1}\ket{0}\ket{1}=\ket{101}$~\cite[p.25-26]{pastorello_concise_2023}. %

We employ a feature map based encoding, as proposed by Havlicek et al.~\cite{havlicek_supervised_2019}. It uses a quantum feature map to transform the features and aims to exploit the dimensionality of the Hilbert space. %

Equation~\ref{eq:pauli_featuremap} shows the so-called Pauli expansion circuit. $i$ refers to the imaginary unit and $n$ to the number of qubits. $P$ is a set of Pauli gates, consisting of $X$, $Y$ (see Equation~\ref{eq:y}), $Z$, and identity gates $\mathbf{I}$. 

\begin{equation}\label{eq:pauli_featuremap}
    U_{\Phi(x)} = exp(i\sum_{S\subseteq [n]}\phi_S(x) \prod_{i\in S}P_i)
\end{equation}

\begin{equation}\label{eq:y}
     Y = \begin{pmatrix} 0 & -i \\ i & 0 \end{pmatrix}
\end{equation}

$S\in\genfrac{()}{}{0pt}{2}{n}{k}$ describes the connections between the qubits. The connections are influenced by $k \in \{1,\dots,n\}$, the entanglement hyperparameter and the Pauli matrices. $\phi_S(x)$ refers to a non-linear function. Equation~\ref{eq:psi} shows the default non-linear function in Qiskit. \footnote{\url{https://qiskit.org/documentation/stubs/qiskit.circuit.library.PauliFeatureMap.html}. Accessed 20.06.2023.}.

\begin{equation}\label{eq:psi}
    \phi_S(x) = \begin{cases}
        x_0, k = 1 \\
        \prod_{j \in S} (\pi - x_j), \text{else}
    \end{cases}
\end{equation}

The tensor product of $n$ Hadamard gates (see Equation~\ref{eq:h}) is then applied to the input qubits (all in state $|0\rangle$), which puts them in an equal superposition of all possible states (if we measure we have the same probability for all $2^n$ states). Afterward, the Pauli expansion circuit is applied. The two operations can be repeated arbitrary times. Equation~\ref{eq:pauli_encoding} shows the final feature map with two repetitions.

\begin{equation}\label{eq:h}
    H = \frac{1}{\sqrt{2}} \begin{pmatrix}
    1 & 1 \\ 1 & -1
\end{pmatrix}
\end{equation}

\begin{equation}\label{eq:pauli_encoding}
    |\Phi(x)\rangle = U_{\Phi(x)}H^{\otimes n} U_{\Phi(x)}H^{\otimes n}|0\rangle_n
\end{equation}

Qiskit provides two variants: ZZFeatureMap and ZFeatureMap. The ZFeatureMap sets $k=1$ and $P_0=Z$, hence not entangling the qubits, and the ZZFeatureMap sets $k=2$, $P_0=Z$ and $P_{0,1} = ZZ$. The implementations use one qubit for every feature present in the data. We leave all hyperparameters of the ZFeatureMap to their default values. We optimize the entanglement strategy for the ZZFeatureMap, which allows the same options as TwoLocal.

\section{Experimental Setup}
\label{sec:experimental}
\subsection{Hardware Specifications}
Quantum simulators have the advantage of being capable of simulating a perfect quantum computer on a classical one without any noise, allowing to create baselines to compare the results to experiments executed on real quantum machines. By incorporating noise models in simulators, one can obtain a more accurate idea of the results on a real quantum computer~\cite{huang_near-term_2023}. 

Thus, we conduct our experiments using quantum simulators, specifically, we employ the simulators provided by Qiskit Aer. We first run the experiments on a perfect simulator, before incorporating noise models generated from real machines. We choose a noise model from IBM's 7-qubit quantum computer called “Perth”\footnote{Available through \url{https://docs.quantum.ibm.com/api/qiskit/qiskit.providers.fake_provider.FakePerth}. Accessed 16.01.2024}.

While quantum simulators have several advantages, e.g., availability and controlled noise, %
phenomena that are inherently quantum (i.e., entanglement) are computationally intense to simulate on a classical computer, which makes comparing runtimes more difficult. Therefore, we analyze and compare the runtime mainly for aspects which can also be expected on real quantum hardware (e.g., number of parameters or circuit depth).

We run our experiments on a Intel(R) Xeon(R) CPU E5-2623 v4 (16 cores @ 2.60GHz) server with 128GB RAM and a Debian/GNU Linux 11 OS. We use Python v3.11.7 and the IBM Qiskit framework v0.43~\cite{Qiskit} for the implementation.

\subsection{Experiments}
\begin{table*}
    \centering
    \begin{tabular}{|c|c|c|c|c|c|c|c|c|c|c|}\hline
         & \#Dataset & Prepr. & Feat. Map & F. Ent. & Ansatz & A. Ent. & Opt. & Init. & Noiseless & Noisy  \\\hline
         \cite{sierra-sosa_data_2023} & 5 & \checkmark & \checkmark &  &  &  &  &  & \checkmark &  \\
         \cite{mancilla_preprocessing_2022} & 2 & \checkmark &  &  &  &  &  &  & \checkmark &  \\
         \cite{hancco-quispe_quantum_2023} & 1 & \checkmark &  &  &  &  &  &  & \checkmark &  \\
         \cite{joshi_evaluating_2021} & 1 &  &  &  &  &  & \checkmark &  &  & \checkmark \\
         \cite{suryotrisongko_hybrid_2022} & 1 &  & \checkmark &  & \checkmark &  & \checkmark &  & \checkmark &  \\
         \cite{piatrenka_quantum_2022} & 1 &  & \checkmark &  & \checkmark &  & \checkmark &  & \checkmark & \checkmark \\
         \cite{katyayan_supervised_2023} & 1 & \checkmark & \checkmark &  & \checkmark  &  &  &  &  & \checkmark \\
         \cite{joshi_comparing_2021} & 1 &  &  &  & \checkmark &  & \checkmark &  &  & \checkmark \\
         \cellcolor[gray]{0.8} \textbf{This work} & \cellcolor[gray]{0.8} \textbf{4} & \cellcolor[gray]{0.8} \checkmark & \cellcolor[gray]{0.8} \checkmark & \cellcolor[gray]{0.8} \checkmark & \cellcolor[gray]{0.8} \checkmark & \cellcolor[gray]{0.8} \checkmark & \cellcolor[gray]{0.8} \checkmark & \cellcolor[gray]{0.8} \checkmark & \cellcolor[gray]{0.8} \checkmark & \cellcolor[gray]{0.8} \checkmark \\\hline
    \end{tabular}
    \caption{Comparison of Studies Investigating QNN Hyperparameters}
    \label{tab:comparison_studies}
\end{table*}

We evaluate all hyperparameter configurations of Section~\ref{benchmarking_methodology},  summing up to 1512 configurations per dataset and noise configuration. We compare our setup to related work in Table~\ref{tab:comparison_studies}.

Concerning optimizers, to assess the convergence behavior and set an appropriate iteration count for every dataset, noise setting and optimizer, we plot the training loss at different iterations. For COBYLA, convergence typically occurs within a range of $250$ to $500$ iterations. The SPSA optimizer converges within $125$ to $300$ iterations, and the Nelder-Mead optimizer within $100$ to $250$ iterations. Additionally, we apply early stopping with a tolerance value of $0.1$ for the loss function, which is supported by COBYLA and Nelder-Mead. %

Moreover, SPSA and Nelder-Mead work by probing solutions (which is done in so-called steps) and only move on to the next iteration if they accept the step. We define the termination criterion based on the number of iterations, however, in the results section, we analyze the runtime of these two optimizers based on the number of steps they took. To avoid any confusion with the reported numbers, we want to actively highlight that the terms iteration and step are not used interchangeably in our analysis.

Furthermore, we activate the adaptive hyperparameter option to improve convergence of Nelder-Mead, which adapts the algorithm's parameters to the problem dimensionality, since the optimizer struggles to find a reasonable path in some scenarios.

Since the runtime increases quickly with more data, we limit the samples used in the experiments to a maximum of $400$ samples for training and $250$ for testing the final models, which limits the runtime to ten hours. %
The trained models are evaluated using the accuracy and weighted f-1 score.

\section{Results}
\label{sec:evaluation}
To better visualize the results, we define the set of best and worst configurations as those that are within 10\% accuracy of the best or worst one overall. Furthermore, we study how noise affects the configurations by running a pairwise analysis, comparing them with and without noise directly.

To ensure statistical significance, we perform non-parametric statistical tests with a significance level of 5\%. In particular, we choose a Friedman Test for dependent distributions of more than two groups and the Wilcoxon Signed Rank test for two dependent distributions. Moreover, we use the Kruskal-Wallis test for independent distributions of more than two groups and the Mann-Whitney-U test for independent distributions of two groups.

In the following, we distinguish between \textit{observations}, which are general patterns we identify and \textit{main findings}, which are consistent patterns that constitute key insights into QNN training.

\subsection{Overview}

\begin{table*}[!th]
    \centering
    \begin{tabular}{|c|c|c|c|c|c|c|c|}\hline
    Acc. & F-1 & Time [s] & Ansatz/Entanglement & Optimizer & Feature Map/Ent. & Prepr. & Init \\\hline
    \multicolumn{8}{|l|}{\textbf{Noiseless}}\\\hline
0.972 &	0.965 &	3848 &	EfficientSU2/sca &	SPSA &	ZZFeatureMap/sca & LDA & Beta \\
0.968 &	0.961 &	3740 &	EfficientSU2/sca &	SPSA &	ZZFeatureMap/linear & LDA & Beta \\
0.964 &	0.960 &	3413 &	EfficientSU2/sca &	COBYLA & ZZFeatureMap/circular & LDA & Beta \\
0.964 &	0.959 &	3022 &	EfficientSU2/sca &	COBYLA & ZZFeatureMap/pairw. & LDA & Beta \\
0.972 &	0.958 &	2598 &	RealAmplitudes/circular & SPSA & ZFeatureMap & LDA & Beta \\\hline
\multicolumn{8}{|l|}{\textbf{Noisy}}\\\hline
0.968 &	0.954 &	7300 &	TwoLocal/linear &	COBYLA & ZFeatureMap & LDA & Beta \\
0.960 &	0.950 &	6462 &	TwoLocal/sca &	COBYLA & ZFeatureMap & LDA & Beta \\
0.964 &	0.948 &	6248 &	TwoLocal/pairw. &	COBYLA & ZFeatureMap & LDA & Beta \\
0.960 &	0.944 &	6947 &	RealAmplitudes/linear &	COBYLA & ZFeatureMap & LDA & Beta \\
0.960 &	0.944 &	8569 &	EfficientSU2/linear &	SPSA & ZFeatureMap & LDA & Beta \\\hline
    \end{tabular}
    \caption{KDD Cup: Top Five Configurations}
    \label{tab:kdd_top5}
\end{table*}

The accuracy for the noiseless and noisy KDD Cup experiments ranges from $0$\%-$97$\%. The top five configurations according to the f-1 score in the different settings are shown in Table{~\ref{tab:kdd_top5}}. We choose the f-1 score as the final criterion, as it represents both Precision and Recall equally, rather than just the ratio of correctly predicted samples. All use LDA for preprocessing and beta distribution initialization. Most of the best configurations in the noiseless setting use EfficientSU2 and ZZFeatureMap. In the noisy one, TwoLocal and ZFeatureMap are more frequent. The mean difference between the configurations with and without noise is at $10$\% accuracy, with a standard deviation of $11$\%. The minimum difference lies at $0$\%, the maximum at $67$\%.

Both results for the noisy and noiseless Cover Type experiments range from $2$\%-$61$\% accuracy. We show the top five configurations from both settings in Table~\ref{tab:covtype_top5}. Again, the noisy configurations are slightly worse, all use LDA for dimensionality reduction and most configurations use beta distribution initialization and ZZFeatureMap in both settings. %

\begin{figure}[!th]
    \centering
    \includegraphics[width=\columnwidth]{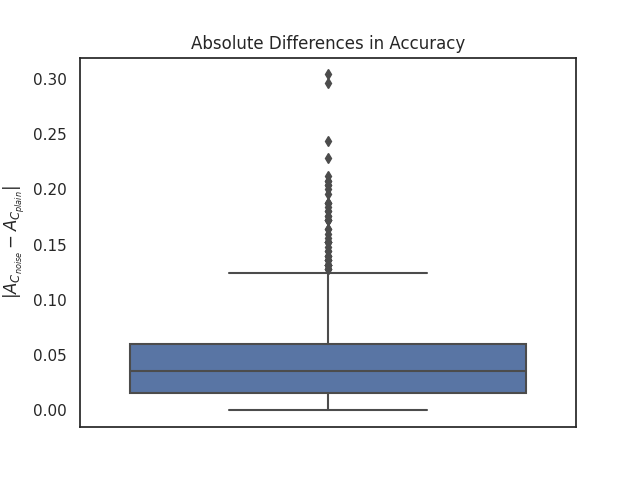}
    \caption{Cover Type: Absolute Difference in Accuracy}
    \label{fig:covtype_abs_diff}
\end{figure}

The mean difference in accuracy between noisy and noiseless configurations is $4$\% with a standard deviation of $3$\%. Figure~\ref{fig:covtype_abs_diff} shows that more than $75$\% of configurations have a difference $<8$\%, the maximum being at $30$\%. %
Noisy simulators lead to significantly better results for the Cover Type dataset. However, the best models in the noiseless setting outperform the best ones in the noisy setting.

\begin{table*}
    \centering
    \begin{tabular}{|c|c|c|c|c|c|c|c|}\hline
            Acc. & F-1 & Time [s]] & Ansatz/Entanglement & Optimizer & Feature Map/Ent. & Prepr. & Init \\\hline
    \multicolumn{8}{|l|}{\textbf{Noiseless}}\\\hline
0.612 & 0.612 & 2066 & EfficientSU2/circular & COBYLA & ZZFeatureMap/linear & LDA & Beta \\
0.608 & 0.612 & 2097 & EfficientSU2/linear & SPSA & ZZFeatureMap/sca	& LDA & Beta \\
0.612 & 0.610 & 886 & PauliTwoDesign/full & COBYLA & ZZFeatureMap/pairw. & LDA & Beta \\
0.616 & 0.609 & 1422 & RealAmpl./linear & SPSA & ZFeatureMap & LDA & Nor. \\
0.612 &	0.605 & 2100 & EfficientSU2/circular & SPSA & ZZFeatureMap/circular & LDA & Beta \\\hline
\multicolumn{8}{|l|}{\textbf{Noisy}}\\\hline
0.608 & 0.600 & 1867 & TwoLocal/linear & SPSA & ZZFeatureMap/pairw. & LDA & Beta \\
0.616 & 0.593 & 2031 & RealAmpl./linear & COBYLA & ZFeatureMap & LDA & Beta \\
0.588 & 0.592 & 2493 & RealAmpl./circular & COBYLA & ZZFeatureMap/circular & LDA & Beta \\
0.596 & 0.592 & 3190 & PauliTwoDesign/full & COBYLA & ZZFeatureMap/linear & LDA & Beta \\
0.596 & 0.588 & 1922 & TwoLocal/full & COBYLA & ZZFeatureMap/linear & LDA & Beta \\\hline
    \end{tabular}
    \caption{Cover Type: Top Five Configurations}
    \label{tab:covtype_top5}
\end{table*}

The results from the Rice dataset are shown in Table \ref{tab:rice_top5}. The noiseless results range from 7\%-91\% accuracy, while the noisy ones range from 9\%-90\%. The configurations with and without noise differ 3\% on average in accuracy, with a standard deviation of 3\%. We again find configurations with exactly the same performance, however, the maximum difference lies at 36\%. Only ZFeatureMap configurations with beta distribution initialization are among the best ones in both settings.

We show the top five configurations with and without noise for the Glass Identification dataset in Table \ref{tab:glass_top5}. Our models had problems fitting the data, with the highest accuracy at only $62$\%, which could be due to the limited samples. The results without noise range from $2$\%-$62$\%, and with noise from $0$\%-$53$\% accuracy. The mean accuracy difference between configurations with and without noise is $7$\%, with a standard deviation of $6$\%. While the minimum difference is at $0$\%, the maximum is at $34$\%. 

\begin{table*}[!th]
    \centering
    \begin{tabular}{|c|c|c|c|c|c|c|c|}\hline
            Acc. & F-1 & Time [s] & Ansatz/Entanglement & Optimizer & Feature Map/Ent. & Prepr. & Init \\\hline
    \multicolumn{8}{|l|}{\textbf{Noiseless}}\\\hline
0.916 & 0.915 & 612 & TwoLocal/pairw. & COBYLA & ZFeatureMap & PCA & Beta \\
0.904 & 0.902 & 1411 & RealAmplitudes/linear & SPSA & ZFeatureMap & PCA & Beta \\
0.900 & 0.899 & 668 & RealAmplitudes/circular & COBYLA & ZFeatureMap & PCA & Beta \\
0.896 & 0.895 & 688 & RealAmplitudes/sca & COBYLA & ZFeatureMap & PCA & Beta \\
0.896 & 0.895 & 8023 & TwoLocal/sca & COBYLA & ZFeatureMap & PCA & Beta \\\hline
\multicolumn{8}{|l|}{\textbf{Noisy}}\\\hline
0.908 & 0.907 & 7470 & PauliTwoDesign & SPSA & ZFeatureMap & PCA & Beta \\
0.896 & 0.895 & 4785 & RealAmplitudes/sca & COBYLA & ZFeatureMap  & PCA & Beta \\
0.896 & 0.895 & 4518 & TwoLocal/pairw. & COBYLA & ZFeatureMap  & PCA & Beta \\
0.896 & 0.894 & 4573 & TwoLocal/sca & COBYLA & ZFeatureMap  & PCA & Beta \\
0.892 & 0.890 & 9214 & EfficientSU2/full & COBYLA & ZFeatureMap & PCA & Beta \\\hline
    \end{tabular}
    \caption{Rice: Top Five Configurations}
    \label{tab:rice_top5}
\end{table*}

\begin{table*}[!th]
    \centering
    \begin{tabular}{|c|c|c|c|c|c|c|c|}\hline
            Acc. & F-1 & Time [s] & Ansatz/Entanglement & Optimizer & Feature Map/Ent. & Prepr. & Init \\\hline
    \multicolumn{8}{|l|}{\textbf{Noiseless}}\\\hline
0.604 & 0.571 & 608 & EfficientSU2/linear & COBYLA	& ZZFeatureMap/sca & PCA & Beta \\
0.627 & 0.561 & 470 & EfficientSU2/sca & COBYLA & ZFeatureMap & LDA & Beta \\
0.558 & 0.546 & 641 & TwoLocal/linear & SPSA & ZZFeatureMap/circular & PCA & Beta \\
0.581 & 0.538 & 302 & EfficientSU2/circular & COBYLA & ZFeatureMap & LDA & Beta \\
0.534 & 0.535 & 663 & RealAmplitudes/sca & SPSA & ZZFeatureMap/circular & PCA & Beta \\\hline
\multicolumn{8}{|l|}{\textbf{Noisy}}\\\hline
0.534 & 0.519 & 342 & RealAmplitudes/sca & COBYLA & ZZFeatureMap/circular & LDA & Beta \\
0.488 & 0.497 & 3590 & EfficientSU2/full & COBYLA & ZZFeatureMap/full & PCA & Beta \\
0.511 & 0.487 & 1959 & TwoLocal/sca & SPSA & ZZFeatureMap/linear & PCA & Beta \\ 
0.511 & 0.471 & 1853 & PauliTwoDesign & COBYLA & ZFeatureMap & PCA & Nor. \\
0.488 & 0.465 & 503 & TwoLocal/pairw. & SPSA & ZFeatureMap & LDA & Nor. \\\hline
    \end{tabular}
    \caption{Glass Identification: Top Five Configurations}
    \label{tab:glass_top5}
\end{table*}
 
\subsection{Initialization}
\noindent\mainfindingbox{\textbf{Main Finding 1:} Beta initialization performs significantly better than both normal and uniform initialization in all settings. We find no significant differences between normal and uniform initialization in most settings.}

Therefore, the lower bounds on the magnitude of the gradient when using the normal distribution for initialization from~\cite{NEURIPS2022_7611a3cb}, did not lead to significantly better results in our experiments. We give a detailed overview of the significance tests in the appendix.

Figure{~\ref{fig:covtype_init}} shows the performance of the different initialization methods for the Cover Type dataset. Our results are therefore consistent with the conjecture from{~\cite{beinit}}, that initializing using the beta distribution can lead to better trainability. Our results extend their work by applying large-scale real-world datasets beyond the Wine and Iris datasets tested in their study.

\begin{figure}[!th]
    \centering
    \includegraphics[width=0.45\textwidth]{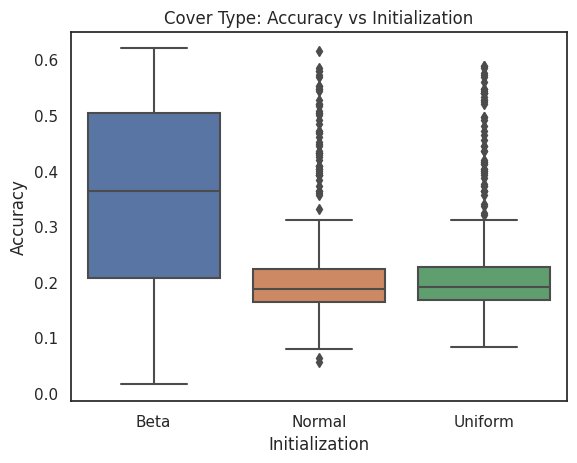}
    \caption{Cover Type, Noiseless: Initialization vs. Accuracy}
    \label{fig:covtype_init}
\end{figure}

Nonetheless, we want to point out that, for the Cover Type dataset, in the noiseless and noisy setting, $8$\% and $5\%$ of the beta initialized configurations, compared to $1$\% and $<1\%$ of uniform and normal configurations, are in the set of worst configurations. Therefore, while leading to significantly better results for some configurations, beta initialization can impact the results negatively as well. However, we find this pattern to be dependent on the dataset, as the pattern for the other datasets is not as clear.

When comparing the noiseless and noisy settings, we find that all initialization methods work significantly better without noise for KDD Cup. Beta initialization works better in the noiseless setting for the Rice dataset, while we find no difference for the other initialization methods and noise settings. Only beta initialization works significantly better without noise for Cover Type and Glass Identification, while uniform leads to better results when noise is introduced. Normal initialization for Glass Identification also works significantly better with noise.

Furthermore, uniform initialization takes significantly less time in the Cover Type noiseless settings, and in all KDD Cup and Glass Identification settings, than beta and normal do. For KDD Cup and Cover Type, we find that this is mainly due to it stopping between 100-450 steps earlier for Nelder-Mead configurations. For Glass Identification, we were not able to pin it down to one factor. 

Normal initialization is significantly faster than beta in all settings except for the Rice dataset, which is explained by COBYLA requiring between $10$ and $50$ more iterations for beta than for normal. We find no significant runtime differences for the noiseless Rice dataset, but uniform initialization takes significantly longer than both others in the noisy setting. We find no clear patterns explaining the behavior, however.

\subsection{Optimizer}
\begin{figure}
    \centering
    \includegraphics[width=0.5\textwidth]{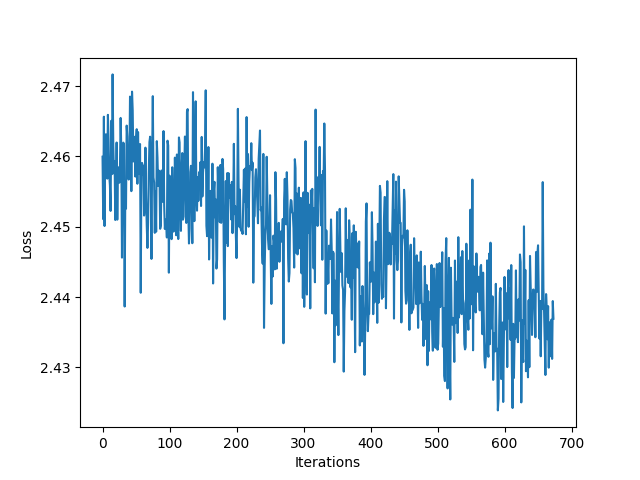}
    \caption{KDD Cup: Nelder-Mead Convergence}
    \label{fig:kdd_nm_conv}
\end{figure}

\noindent\mainfindingbox{\textbf{Main Finding 2:} SPSA and COBYLA significantly outperform Nelder-Mead in all settings. We find no consistent performance patterns for COBYLA and SPSA, but COBYLA takes significantly less time in all settings.}
Nelder-Mead fails to provide a reasonable optimization path, which can be seen from convergence plots such as the one shown in Figure~\ref{fig:kdd_nm_conv}. The only time we find Nelder-Mead in the set of best configurations is in the Glass Identification noisy setting. 

While the algorithm is commonly used, there are relatively few theoretical studies on convergence~\cite{Alexandru_2020}. Several inefficiencies and pitfalls are known, however, the authors in~\cite{nm_conv} identify several reasons why it is still commonly used. First, it is a fairly simple algorithm and, secondly, improves very quickly over current solutions even though it may not converge to optima. Finally, if the method works, it requires few function evaluations, which is particularly relevant for compute-expensive cost functions, where the derivatives are difficult or even impossible to compute.

\begin{figure}
    \centering
    \includegraphics[width=0.45\textwidth]{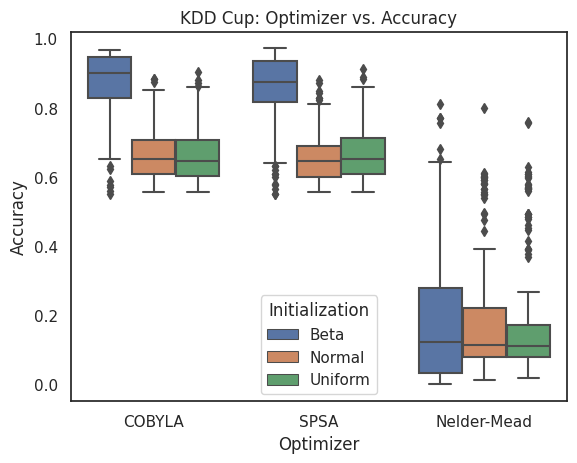}
    \caption{KDD Cup, Noiseless: Optimizer vs. Accuracy}
    \label{fig:kdd_opt}
\end{figure}

COBYLA is significantly better than SPSA for the KDD Cup noiseless setting, whereas SPSA significantly outperforms COBYLA on all Cover Type and the noiseless Glass Identification settings. We find no significant difference between the two optimizers for the noisy KDD Cup and Glass Identification and all Rice settings. Nonetheless, in all cases, they are equally represented among the set of best configurations. 

Figure~\ref{fig:kdd_opt} visualizes the performance of the different optimizers for the noiseless experiments on the KDD Cup dataset. Beta initialization leads to significantly better results for SPSA and COBYLA, however not for Nelder-Mead, suggesting that it does not lead to a position in the loss landscape that allows a better optimization path irrespective of the optimizer.

All optimizers are significantly better in the noiseless KDD Cup and Rice settings than in the noisy ones. For the Cover Type dataset, COBYLA and Nelder-Mead perform significantly better in the noisy setting, in contrast to SPSA, which performs better in the noiseless one. We find no significant differences for COBYLA, SPSA to perform better without noise, and Nelder-Mead to perform significantly better with noise for the Glass Identification dataset.

COBYLA is significantly faster than SPSA and Nelder-Mead in all settings, as is SPSA compared to Nelder-Mead.

\subsection{Ansatz}
\noindent\observationbox{\textbf{Observation:} There are no significant differences between the ansatzes for any dataset, neither in the noiseless nor noisy setting. However, PauliTwoDesign performs slightly worse in mean and is usually less represented among the best configurations.}

In particular, while mean and standard deviation are very similar across all ansatzes, PauliTwoDesign has less extreme minima and maxima, exhibiting a more stable performance when compared to the other ansatzes: for the Cover Type noisy setting, the minimum is $5$\% accuracy, compared to $1$\%-$2$\% for the other ansatzes, and for the KDD Cup noisy setting, the maximum is $89$\%, compared to $96$\%. The same can be observed for the noiseless Glass Identification dataset, with a minimum of $6$\%, compared to $2$\% accuracy for the others, and a maximum of $51$\%, compared to $53$\%-$62$\%. These patterns are not ever-present, but we observe a tendency in our experiments.

When we directly compare noisy and noiseless settings, we find that all ansatzes are significantly better without noise for the KDD Cup and Rice datasets. We find no differences for EfficientSU2 on Cover Type, PauliTwoDesign to perform better without noise, and RealAmplitudes and TwoLocal to lead to better results in the noisy setting. For the Glass Identification dataset, we find no differences for EfficientSU2 and TwoLocal, and RealAmplitudes and PauliTwoDesign to perform significantly better without noise.

\begin{figure}[!th]
    \centering
    \includegraphics[width=0.45\textwidth]{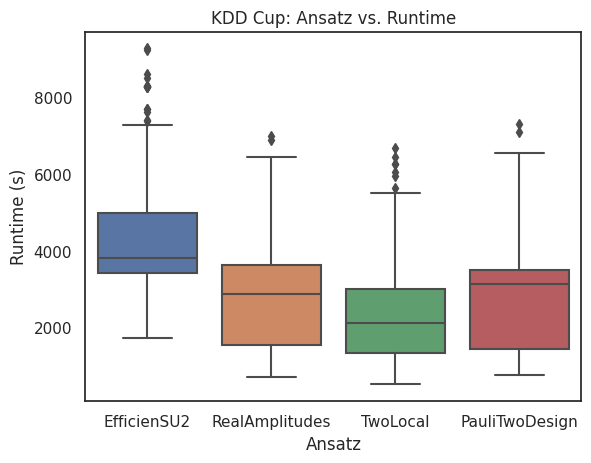}
    \caption{KDD Cup, Noiseless: Ansatz vs. Runtime}
    \label{fig:kdd_noiseless_ansatz_runtime}
\end{figure}

All ansatzes are significantly faster than EfficientSU2. The longer runtime of EfficientSU2 is explained by the default configuration employing more rotation gates than the other ansatzes, resulting in twice the amount of trainable parameters. Our results imply that these additional parameters do not necessarily lead to better performance. Therefore, it may be favorable to stick to the less parametrized and, hence, faster ansatzes. We visualize the runtimes in Figure~\ref{fig:kdd_noiseless_ansatz_runtime}.

\subsection{Ansatz Entanglement}
\noindent\observationbox{\textbf{Observation:} We find no consistent patterns regarding the performance of the different ansatz entanglement strategies.}

For KDD Cup and Cover Type, neither the noiseless nor noisy experiments show any significant differences for the ansatz entanglement strategies. When we compare the noisy and noiseless settings directly, we find no significant differences either. We find some significant differences for the Glass Identification and Rice dataset, however, no consistent patterns. For completeness, we list them in the appendix.

\subsection{Feature Map}
\noindent\observationbox{\textbf{Observation:} Beta initialization boosts the performance of ZZFeatureMap configurations in most settings significantly. Still, the performance of the feature maps is dependent on the dataset.}

ZFeatureMap is significantly better than ZZFeatureMap in the KDD Cup noiseless, and in all Cover Type and Glass Identification settings. Still, we find that the majority of best configurations use ZZFeatureMap in these settings. This sounds counterintuitive at first, however, beta initialization boosts the performance of ZZFeatureMap configurations by $12$\%-$16$\% accuracy in mean, compared to $2$\%-$5$\% for ZFeatureMap. Hence, ZZFeatureMap configurations are significantly worse without beta initialization than ZFeatureMap ones, explaining the significance tests. Still, beta initialization boosts the performance of ZZFeatureMap configurations to make up the set of best configurations.

\begin{figure}[!th]
    \centering
    \includegraphics[width=0.45\textwidth]{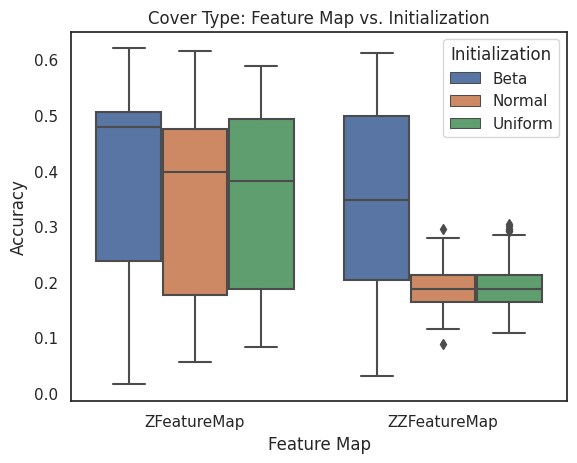}
    \caption{Cover Type, Noiseless: Feature Map vs. Initialization}
    \label{fig:covtype_noiseless_featmap_init}
\end{figure}

Entanglement in a circuit, as is used in ZZFeatureMap compared to ZFeatureMap, has been linked to BPs~\cite{entanglement_induced}. Our results suggest that the proposed beta initialization helps to cope with them, as the performance difference to normal and uniform is significant. We visualize the different feature maps and initializations in Figure~\ref{fig:covtype_noiseless_featmap_init}. 

\begin{figure}
    \centering
    \includegraphics[width=0.45\textwidth]{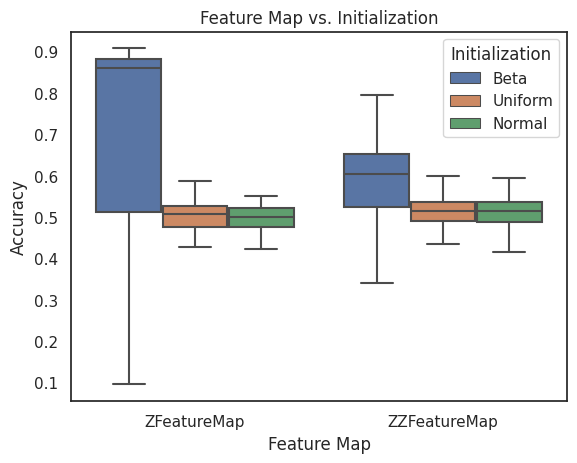}
    \caption{Rice, Noisy: Feature Map vs. Initialization}
    \label{fig:rice_noise_featmap_init}
\end{figure}

In contrast, there are no significant differences between the feature maps in the KDD Cup noisy setting and all Rice settings. Here, the very best configurations are ZFeatureMap ones. For KDD, we find that beta initialization does not boost the performance of ZZFeatureMap as much. Interestingly, for the Rice dataset, beta initialization boosts ZFeatureMap configurations a lot more ($19$\%-$22$\% accuracy in mean) than ZZFeatureMap ones ($7$\%-$9$\%), i.e., the behavior is reversed (see Figure~\ref{fig:rice_noise_featmap_init}).

When comparing noisy and noiseless settings, we find no consistent patterns. Considering runtime, ZZFeatureMap configurations take significantly longer than ZFeatureMap ones, since the non-linearity and entanglement between the qubits strongly increase the computational complexity of feature encoding.

\subsection{Feature Map Entanglement}

\noindent\observationbox{\textbf{Observation:} Similarly to the ansatz entanglement, there are no consistent significant differences between the strategies.}

We again find significant differences in some experiments, however, no consistent patterns. Still, full entanglement is usually underrepresented among the best configurations. We again refer to the appendix for further information.

We find no consistent significant differences when directly comparing noisy and noiseless settings.

\subsection{Preprocessing}
\noindent\observationbox{\textbf{Observation:} LDA works significantly better than PCA in most settings or at least leads to a majority of best configurations. This suggests that it may transform the loss landscape in a way that aids optimization.}

While PCA outperforms LDA in the KDD noiseless and noisy settings, about $60$\% of the best configurations use LDA as a preprocessing step in both settings. LDA significantly outperforms PCA for the Cover Type dataset, with $98$\% of the best configurations using LDA in both the noisy and noiseless setting. Furthermore, we find LDA to perform significantly better in the Glass Identification noiseless (1\% accuracy in mean) and noisy (4\%) setting. We observe a tendency that the choice for this hyperparameter depends on the dataset, which we elaborate on in the appendix.

Furthermore, in Figure{~\ref{fig:kdd_noiseless_prepr_init}}, we observe that beta initialization boosts the performance for LDA more than for PCA. The pattern cannot be observed for the Glass Identification dataset, where beta initialization boosts the performance of both preprocessing techniques by $10$\%-$11$\% in mean for the noiseless and by 5\%-6\% in the noisy setting.

LDA arranges the features in a way that better separates the different classes, whereas PCA focuses on keeping the variance in the data. Beta initialization boosts the mean performance of PCA configurations for the KDD Cup and Cover Type datasets about $10$\% in our experiments, compared to $20$\% for LDA. Our results therefore suggest that LDA can transform the loss landscape such that, given a good starting point, the model is likely to end up in a better solution than with PCA.

When we directly compare the configurations with and without noise, we find that PCA and LDA are significantly better without noise for the KDD Cup dataset (5\% difference in accuracy in mean). For the Cover Type dataset, we find no significant differences for LDA, but PCA works significantly better with noise (1\% in mean). For Glass Identification, we find no significant differences for PCA, but LDA works significantly better with noise (2\%).

When considering the runtime, we find no significant differences for the KDD Cup dataset, but PCA configurations run significantly longer than LDA ones for the Cover Type and Glass Identification datasets. We find that COBYLA PCA configurations take more iterations in mean than LDA ones, as do Nelder-Mead ones, explaining the runtime difference. Furthermore, the significantly longer convergence could also indicate that the loss landscape of PCA configurations can be more difficult to navigate than LDA ones.

\begin{figure}[!th]
    \centering
    \includegraphics[width=0.45\textwidth]{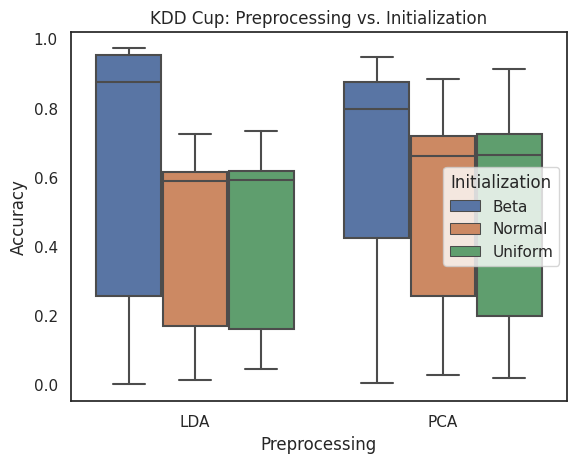}
    \caption{KDD Cup, Noiseless: Preprocessing vs. Initialization}
    \label{fig:kdd_noiseless_prepr_init}
\end{figure}

\subsection{Further Analysis}
One of the most prevalent patterns we discover is that the optimizer is the most crucial hyperparameter to set. Using Nelder-Mead in our experiments consistently led to underperforming configurations, independently of any other hyperparameters. 

\begin{figure}[!th]
    \centering
    \includegraphics[width=0.45\textwidth]{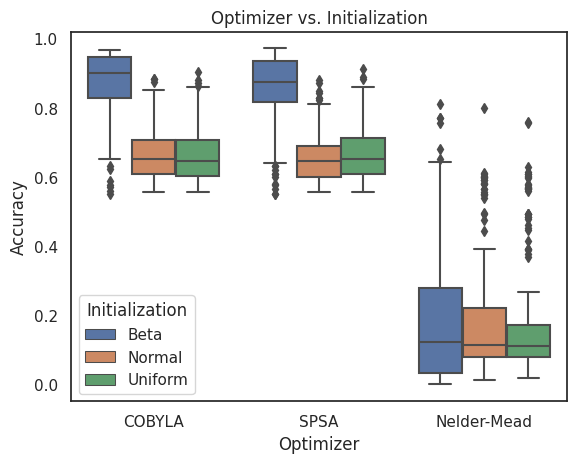}
    \caption{KDD Cup, Noisy: Optimizer vs. Initialization}
    \label{fig:kdd_init_opt}
\end{figure}

Furthermore, our results show that the initialization strategy plays a crucial role. Nonetheless, we find that the initialization strategy does not necessarily help all optimizers. Figure{~\ref{fig:kdd_init_opt}} shows that even when combining initialization and optimizer, the optimizer is the most influential parameter for a model's performance.

\section{Discussion}
\label{sec:discussion}

\begin{table*}[!th]
    \centering
    \begin{tabular}{|c|cccc|cccc|cccc|}\hline
         & \multicolumn{4}{c|}{Beta} & \multicolumn{4}{c|}{Normal} & \multicolumn{4}{c|}{Static} \\\hline
         & K & C & G & R & K & C & G & R & K & C & G & R  \\ \hline
        Beta &  &  &  & & \cellcolor{yellow}0.35 & \cellcolor{red}7e-4 & \cellcolor{yellow}0.84 & \cellcolor{yellow}0.54 &  \cellcolor{yellow}0.45 & \cellcolor{yellow}0.55 & \cellcolor{yellow}0.81 & \cellcolor{yellow}0.22  \\\hline
        Normal & \cellcolor{yellow}0.35 & \cellcolor{green}7e-4 & \cellcolor{yellow}0.84 & \cellcolor{yellow}0.54 & & &  & & \cellcolor{yellow}0.69 & \cellcolor{green}7e-5 & \cellcolor{yellow}0.81 & \cellcolor{yellow}0.12 \\\hline
        Static & \cellcolor{yellow}0.45 & \cellcolor{yellow}0.55 & \cellcolor{yellow}0.81 & \cellcolor{yellow}0.22 & \cellcolor{yellow}0.69 & \cellcolor{red}7e-5 & \cellcolor{yellow}0.81 & \cellcolor{yellow}0.12 & & & &  \\\hline
    \end{tabular}
    \caption{Initialization with Beta Distribution Mean and SD: Significance Tests}
    \label{tab:init_mean_sd}
\end{table*}

Our experiments gave us valuable insights into tuning hyperparameters for QNNs. The empirical evidence we gathered suggests that the choice of optimizer and %
initialization method is crucial. In particular, COBYLA and SPSA seem to both be reasonable choices for the optimizer. Since COBYLA configurations are usually faster, we suggest using it as a starting point. %
Furthermore, it is favorable to use beta initialization compared to random and normal one, in particular when employing an entangled feature map. 

Although ZFeatureMap was often significantly better than ZZFeatureMap, in most settings, the best configurations use ZZFeatureMap. Based on the significant performance differences between beta initialized and non-beta initialized configurations for ZZFeatureMap compared to ZFeatureMap, we hypothesize that entanglement-induced barren plateaus occur and that a clever initialization strategy can indeed help in finding better optimization trajectories.

\noindent\mainfindingbox{\textbf{Main Finding 3:} While beta distribution initialization works significantly better than all others in our main experiments, post-hoc tests reveal that it is not the probability distribution, but rather the range of values used, that is indicative of the likelihood of encountering a BP. This stresses the importance of exploring the theoretical aspects of parameter initialization further.}
After looking at the significant results, we were curious about why exactly beta distribution initialization works so much better than the other two. Therefore, we ran two additional post-hoc tests, initializing all parameters to the mean of the beta distribution, and using a normal distribution with the mean and standard deviation of the chosen beta distribution. We hardly found any significant differences between the three techniques, which we visualize in Table~\ref{tab:init_mean_sd}. Red/green means the row is significantly worse/better than the column on the specific dataset (K: KDD Cup, C: Cover Type, G: Glass Identification, R: Rice), whereas yellow denotes no significant differences. We test using a Wilcoxon test with a significance level of 0.05 and the numbers denote the p-value. %

Furthermore, to our knowledge, there are no indicators of a causal relationship between the distribution of the data and the parameters determining the rotations applied in the circuit (as is assumed in~\cite{beinit}). Therefore, it is an open question whether certain value ranges just lead to better initial positions, or there indeed is a connection between the two, which would be important to study from a theoretical point of view.

The ansatzes perform similarly on average, however, some are more likely to lead to higher-performing outliers. Therefore, we advise starting with RealAmplitudes, which can be used out-of-the-box and takes less time than EfficientSU2. Also, TwoLocal is a good option, however, the entanglement and rotation gates have to be set accordingly. We found hardly any significant differences between entanglement strategies, therefore believe using the default option is a good start.

We find that the entanglement strategy has a negligible effect on the performance of the model. We hypothesize that entangling the qubits is indeed important to exploit quantum effects, however, the actual way this is done does not seem to have an impact on the performance of the models. 

For the KDD Cup and Cover Type dataset, using beta initialization in combination with LDA significantly boosts the performance of the models, therefore, all the best models use LDA. We hypothesize that, as LDA separates the classes during preprocessing already, this, together with advantageous initial starting points from beta initialization, can lead to better trainability of the model.

\subsection{Threats to Validity}
\label{sec:threats}

\textit{Scope of the experiment:} While the experiments on the four datasets allowed us to draw valuable conclusions, the results could vary with different ones.

\textit{Reduced dataset:} It is a well-known fact that ML models profit from more data. As we had to limit the training samples, the results could vary when rerunning the experiments with the whole dataset.

\textit{Scalability:} NISQ hardware and the complexity of simulating quantum computers limit today's experiments. It remains an open research question what happens when the experiments are scaled from tens to hundreds or thousands of qubits. 

\textit{Architecture:} We chose the general-purpose QASM simulator~\cite{qasm} for our experiments, however, there are few studies comparing the performance of different simulators. Therefore, the results may differ when using a different one.

\section{Conclusion and Future Work}
\label{sec:conclusion}
In this work, we have emphasized %
the potential of QML in addressing the challenges of Post-Moore era. However, the complexity of the models, combined with few hyperparameter tuning studies with only limited scope to this date, hinder the adoption and further exploration. Therefore, we have collected data about the performance of QML models on various datasets using different hyperparameter configurations with and without noise and evaluated our results rigorously. Our results support researchers and practitioners who want to explore QML and potential applications further, by providing starting points for tuning the models.

Besides collecting data, our experiments on real-world datasets provided us with evidence that initializing using the Gaussian distribution, as proposed in~\cite{NEURIPS2022_7611a3cb}, does not lead to advantages over initializing using a uniform distribution in training the models. Furthermore, in contrast to the conjecture from~\cite{beinit}, we found in additional experiments that the beta distribution itself does not necessarily lead to better initial points and therefore to better predictive performance of the models. Indeed, we mostly found no significant differences when initializing using a normal distribution with the same mean and standard deviation as the beta distribution, and initializing all parameters to the mean of the beta distribution.

Nonetheless, given the importance of theoretically studying these models further, in particular with respect to trainability issues such as BPs, we plan to further explore the model space and loss landscapes. While it seems that clever initialization can mitigate BPs to some extent, we want to consider why certain initial parameter ranges result in better models overall. Also, we plan to extend this study by considering different types of datasets and different types of quantum architectures.

\section*{Acknowledgements}
This work has been partially funded through the Rucon project (Runtime Control in Multi Clouds), Austrian Science Fund (FWF): Y904-N31 START-Programm 2015, by the CHIST-ERA grant CHIST-ERA-19-CES-005, Austrian Science Fund (FWF), Standalone Project Transprecise Edge Computing (Triton), Austrian Science Fund (FWF): P 36870-N, and by Flagship Project HPQC (High Performance Integrated Quantum Computing) \# 897481 Austrian Research Promotion Agency (FFG). We acknowledge the use of IBM Quantum services for this work. The views expressed are those of the authors, and do not reflect the official policy or position of IBM or the IBM Quantum team.

\bibliographystyle{plain}
\bibliography{quantum-ml}

\appendix
\section*{Appendix}
In the following, we will elaborate further on the results of the significance tests we conducted. Red/green means that the row is significantly worse/better than the column. yellow denotes no significant differences and the values are the p-values, rounded to the fourth decimal.

\subsection*{Initialization}
Table~\ref{tab:init_sign} shows the comparison of the three main initialization strategies we tested. We want to highlight that beta initialization was significantly better in \textit{all} experiments conducted than all other strategies. Furthermore, we find significant differences between normal and uniform initialization in only one setting.

\begin{table*}
    \centering
    \begin{tabular}{|c|c|cccc|cccc|cccc|}\hline
         & & \multicolumn{4}{c|}{Beta} & \multicolumn{4}{c|}{Normal} & \multicolumn{4}{c|}{Uniform} \\\hline
         & & K & C & G & R & K & C & G & R & K & C & G & R  \\ \hline
        \multirow[c]{2}{*}[0in]{Beta} & P &  &  &  & & \cellcolor{green}0.0000 & \cellcolor{green}0.0000 & \cellcolor{green}0.0000 & \cellcolor{green}0.0000 &  \cellcolor{green}0.0000 & \cellcolor{green}0.0000 & \cellcolor{green}0.0000 & \cellcolor{green}0.0000  \\
         & N &  &  &  & & \cellcolor{green}0.0000 & \cellcolor{green}0.0000 & \cellcolor{green}0.0000 & \cellcolor{green}0.0000 &  \cellcolor{green}0.0000 & \cellcolor{green}0.0000 & \cellcolor{green}0.0000 & \cellcolor{green}0.0000  \\\hline
        \multirow[c]{2}{*}[0in]{Normal} & P & \cellcolor{red}0.0000 & \cellcolor{red}0.0000 & \cellcolor{red}0.0000 & \cellcolor{red}0.0000 & & &  & & \cellcolor{yellow}0.3375 & \cellcolor{yellow}0.2777 & \cellcolor{yellow}0.3385 & \cellcolor{yellow}0.4354 \\
        & N & \cellcolor{red}0.0000 & \cellcolor{red}0.0000 & \cellcolor{red}0.0000 & \cellcolor{red}0.0000 &  & & & & \cellcolor{red}0.0000 & \cellcolor{yellow}0.5102 & \cellcolor{yellow}0.6722 & \cellcolor{yellow}0.4412 \\\hline
        \multirow[c]{2}{*}[0in]{Uniform} & P & \cellcolor{red}0.0000 & \cellcolor{red}0.0000 & \cellcolor{red}0.0000 & \cellcolor{red}0.0000 & \cellcolor{yellow}0.3375 & \cellcolor{yellow}0.2777 & \cellcolor{yellow}0.3385 & \cellcolor{yellow}0.4354 & & & &  \\
        & N  & \cellcolor{red}0.0000 & \cellcolor{red}0.0000 & \cellcolor{red}0.0000 & \cellcolor{red}0.0000 & \cellcolor{green}0.0000 & \cellcolor{yellow}0.5102 & \cellcolor{yellow}0.6722 & \cellcolor{yellow}0.4412 &  & &  &  \\\hline
    \end{tabular}
    \caption{Initialization: Significance Tests}
    \label{tab:init_sign}
\end{table*}

\subsection*{Preprocessing}
Table~\ref{tab:prepr_sign} compares the two preprocessing techniques. In particular, we would like to highlight that classical preprocessing, in our experiments, has a significant impact on the results. We find significant differences in \textit{all} experiments, and the results always hold for the noisy and noiseless experiments. This suggests that the dataset is an important factor.

\begin{table*}
    \centering
    \begin{tabular}{|c|c|ccc|ccc|}
         \hline
 & & \multicolumn{3}{c|}{LDA} & \multicolumn{3}{c|}{PCA} \\
 & & K & C & G & K & C & G \\\hline
\multirow[c]{2}{*}{LDA} & P &   &   &   & \cellcolor{red}0.0000 & \cellcolor{green}0.0000 & \cellcolor{green}0.0008 \\
 & N &   &   &   & \cellcolor{red}0.0000 & \cellcolor{green}0.0000 & \cellcolor{green}0.0000 \\\hline
\multirow[c]{2}{*}{PCA} & P & \cellcolor{green}0.0000 & \cellcolor{red}0.0000 & \cellcolor{red}0.0008 &   &   &   \\
 & N & \cellcolor{green}0.0000 & \cellcolor{red}0.0000 & \cellcolor{red}0.0000 &   &   &   \\
\hline
    \end{tabular}
    \caption{Preprocessing: Significance Tests}
    \label{tab:prepr_sign}
\end{table*}

\subsection*{Feature Map Entanglement}
Table~\ref{tab:fmap_ent_init1} compares the feature map entanglement. We do not find patterns that are ever-present, however, full entanglement is never significantly better than any other entanglement strategy and is often outperformed. This stands in contrast to linear and pairwise entanglement, which are both never outperformed by any other strategy.

\begin{table*}
    \centering
    \begin{tabular}{|c|c|cccc|cccc|cccc|}
\hline
 & & \multicolumn{4}{c|}{circular} & \multicolumn{4}{c|}{full} & \multicolumn{4}{c|}{linear} \\
 & & K & C & G & R & K & C & G & R & K & C & G & R \\
\hline
\multirow[t]{2}{*}{circular} & P &   &   &   &   & \cellcolor{green}0.0001 & \cellcolor{green}0.018 & \cellcolor{yellow}0.2119 & \cellcolor{yellow}0.7301 & \cellcolor{yellow}0.1493 & \cellcolor{red}0.0000 & \cellcolor{yellow}0.4048 & \cellcolor{yellow}0.5988 \\
 & N &   &   &   &   & \cellcolor{green}0.0000 & \cellcolor{yellow}0.5443 & \cellcolor{yellow}0.2441 & \cellcolor{yellow}0.1297 & \cellcolor{yellow}0.1537 & \cellcolor{red}0.0000 & \cellcolor{yellow}0.4214 & \cellcolor{yellow}0.8146 \\\hline
\multirow[t]{2}{*}{full} & P & \cellcolor{red}0.0001 & \cellcolor{red}0.018 & \cellcolor{yellow}0.2119 & \cellcolor{yellow}0.7301 &   &   &   &   & \cellcolor{red}0.0000 & \cellcolor{red}0.0000 & \cellcolor{red}0.0033 & \cellcolor{yellow}0.5714 \\
 & N & \cellcolor{red}0.0000 & \cellcolor{yellow}0.5443 & \cellcolor{yellow}0.2441 & \cellcolor{yellow}0.1297 &   &   &   &   & \cellcolor{red}0.0000 & \cellcolor{red}0.0000 & \cellcolor{yellow}0.6002 & \cellcolor{yellow}0.1949 \\\hline
\multirow[t]{2}{*}{linear} & P & \cellcolor{yellow}0.1493 & \cellcolor{green}0.0000 & \cellcolor{yellow}0.4048 & \cellcolor{yellow}0.5988 & \cellcolor{green}0.0000 & \cellcolor{green}0.0000 & \cellcolor{green}0.0033 & \cellcolor{yellow}0.5714 &   &   &   &   \\
 & N & \cellcolor{yellow}0.1537 & \cellcolor{green}0.0000 & \cellcolor{yellow}0.4214 & \cellcolor{yellow}0.8146 & \cellcolor{green}0.0000 & \cellcolor{green}0.0000 & \cellcolor{yellow}0.6002 & \cellcolor{yellow}0.1949 &   &   &   &   \\\hline
\multirow[t]{2}{*}{pairwise} & P & \cellcolor{yellow}0.0503 & \cellcolor{green}0.0000 & \cellcolor{yellow}0.9165 & \cellcolor{yellow}0.1658 & \cellcolor{green}0.0000 & \cellcolor{green}0.0000 & \cellcolor{yellow}0.1663 & \cellcolor{yellow}0.4471 & \cellcolor{yellow}0.7099 & \cellcolor{yellow}0.2139 & \cellcolor{yellow}0.4942 & \cellcolor{yellow}0.456 \\
 & N & \cellcolor{yellow}0.5574 & \cellcolor{green}0.0000 & \cellcolor{yellow}0.4476 & \cellcolor{yellow}0.7883 & \cellcolor{green}0.0000 & \cellcolor{green}0.0000 & \cellcolor{yellow}0.8776 & \cellcolor{yellow}0.229 & \cellcolor{yellow}0.8585 & \cellcolor{yellow}0.792 & \cellcolor{yellow}0.7253 & \cellcolor{yellow}0.3521 \\\hline
\multirow[t]{2}{*}{sca} & P & \cellcolor{yellow}0.7064 & \cellcolor{yellow}0.3664 & \cellcolor{yellow}0.422 & \cellcolor{yellow}0.6358 & \cellcolor{green}0.0001 & \cellcolor{green}0.0012 & \cellcolor{yellow}0.3264 & \cellcolor{yellow}0.8701 & \cellcolor{yellow}0.2298 & \cellcolor{red}0.0000 & \cellcolor{yellow}0.1139 & \cellcolor{yellow}0.5581 \\
 & N & \cellcolor{yellow}0.3778 & \cellcolor{yellow}0.4753 & \cellcolor{yellow}0.3436 & \cellcolor{yellow}0.8693 & \cellcolor{green}0.0000 & \cellcolor{yellow}0.1038 & \cellcolor{yellow}0.4568 & \cellcolor{yellow}0.0642 & \cellcolor{red}0.023 & \cellcolor{red}0.0001 & \cellcolor{yellow}0.8478 & \cellcolor{yellow}0.8006 \\\hline
 & & \multicolumn{4}{c|}{pairwise} & \multicolumn{4}{c|}{sca} \\\hhline{----------}
\multirow[t]{2}{*}{circular} & P & \cellcolor{yellow}0.0503 & \cellcolor{red}0.0000 & \cellcolor{yellow}0.9165 & \cellcolor{yellow}0.1658 & \cellcolor{yellow}0.7064 & \cellcolor{yellow}0.3664 & \cellcolor{yellow}0.422 & \cellcolor{yellow}0.6358  \\
 & N & \cellcolor{yellow}0.5574 & \cellcolor{red}0.0000 & \cellcolor{yellow}0.4476 & \cellcolor{yellow}0.7883 & \cellcolor{yellow}0.3778 & \cellcolor{yellow}0.4753 & \cellcolor{yellow}0.3436 & \cellcolor{yellow}0.8693 \\\hhline{----------}
\multirow[t]{2}{*}{full} & P & \cellcolor{red}0.0000 & \cellcolor{red}0.0000 & \cellcolor{yellow}0.1663 & \cellcolor{yellow}0.4471 & \cellcolor{red}0.0001 & \cellcolor{red}0.0012 & \cellcolor{yellow}0.3264 & \cellcolor{yellow}0.8701 \\
 & N & \cellcolor{red}0.0000 & \cellcolor{red}0.0000 & \cellcolor{yellow}0.8776 & \cellcolor{yellow}0.229 & \cellcolor{red}0.0000 & \cellcolor{yellow}0.1038 & \cellcolor{yellow}0.4568 & \cellcolor{yellow}0.0642 \\\hhline{----------}
\multirow[t]{2}{*}{linear} & P & \cellcolor{yellow}0.7099 & \cellcolor{yellow}0.2139 & \cellcolor{yellow}0.4942 & \cellcolor{yellow}0.456 & \cellcolor{yellow}0.2298 & \cellcolor{green}0.0000 & \cellcolor{yellow}0.1139 & \cellcolor{yellow}0.5581 \\
 & N & \cellcolor{yellow}0.8585 & \cellcolor{yellow}0.792 & \cellcolor{yellow}0.7253 & \cellcolor{yellow}0.3521 & \cellcolor{green}0.023 & \cellcolor{green}0.0001 & \cellcolor{yellow}0.8478 & \cellcolor{yellow}0.8006 \\\hhline{----------}
\multirow[t]{2}{*}{pairwise} & P &   &   &   &   & \cellcolor{green}0.0093 & \cellcolor{green}0.0000 & \cellcolor{yellow}0.3395 & \cellcolor{yellow}0.3111 \\
 & N &   &   &   &   & \cellcolor{yellow}0.2028 & \cellcolor{green}0.0001 & \cellcolor{yellow}0.7977 & \cellcolor{yellow}0.8099 \\\hhline{----------}
\multirow[t]{2}{*}{sca} & P & \cellcolor{red}0.0093 & \cellcolor{red}0.0000 & \cellcolor{yellow}0.3395 & \cellcolor{yellow}0.3111 &   &   &   &\\
 & N & \cellcolor{yellow}0.2028 & \cellcolor{red}0.0001 & \cellcolor{yellow}0.7977 & \cellcolor{yellow}0.8099 &   &   &   & \\\hhline{----------}
\end{tabular}
    \caption{Feature Map Entanglement: Significance Tests}
    \label{tab:fmap_ent_init1}
\end{table*}

\subsection*{Ansatz Entanglement}
Table~\ref{tab:ans_ent_sign1} compares the ansatz entanglement strategies. Interestingly, we see the tendency of full entanglement being worse than others from the feature map entanglement reversed, i.e., it is either significantly better or not different from others. Furthermore, sca entanglement never provided any advantages in our experiments. 

\begin{table*}
    \centering
    \begin{tabular}{|c|c|cccc|cccc|cccc|}
\hline
 & & \multicolumn{4}{c|}{circular} & \multicolumn{4}{c|}{full} & \multicolumn{4}{c|}{linear} \\
 & & K & C & G & R & K & C & G & R & K & C & G & R \\
\hline
\multirow[c]{2}{*}{circular} & P &   &   &   &   & \cellcolor{yellow}0.7625 & \cellcolor{yellow}0.1441 & \cellcolor{yellow}0.0594 & \cellcolor{red}0.0042 & \cellcolor{yellow}0.9804 & \cellcolor{yellow}0.7872 & \cellcolor{yellow}0.0623 & \cellcolor{yellow}0.2437 \\
 & N &   &   &   &   & \cellcolor{yellow}0.9271 & \cellcolor{yellow}0.2133 & \cellcolor{yellow}0.6949 & \cellcolor{red}0.0000 & \cellcolor{yellow}0.4918 & \cellcolor{yellow}0.3285 & \cellcolor{yellow}0.5024 & \cellcolor{red}0.0085 \\\hline
\multirow[c]{2}{*}{full} & P & \cellcolor{yellow}0.7625 & \cellcolor{yellow}0.1441 & \cellcolor{yellow}0.0594 & \cellcolor{green}0.0042 &   &   &   &   & \cellcolor{yellow}0.8075 & \cellcolor{yellow}0.0922 & \cellcolor{yellow}0.8753 & \cellcolor{yellow}0.1513 \\
 & N & \cellcolor{yellow}0.9271 & \cellcolor{yellow}0.2133 & \cellcolor{yellow}0.6949 & \cellcolor{green}0.0000 &   &   &   &   & \cellcolor{yellow}0.4153 & \cellcolor{green}0.0143 & \cellcolor{yellow}0.291 & \cellcolor{green}0.0243 \\\hline
\multirow[c]{2}{*}{linear} & P & \cellcolor{yellow}0.9804 & \cellcolor{yellow}0.7872 & \cellcolor{yellow}0.0623 & \cellcolor{yellow}0.2437 & \cellcolor{yellow}0.8075 & \cellcolor{yellow}0.0922 & \cellcolor{yellow}0.8753 & \cellcolor{yellow}0.1513 &   &   &   &   \\
 & N & \cellcolor{yellow}0.4918 & \cellcolor{yellow}0.3285 & \cellcolor{yellow}0.5024 & \cellcolor{green}0.0085 & \cellcolor{yellow}0.4153 & \cellcolor{red}0.0143 & \cellcolor{yellow}0.291 & \cellcolor{red}0.0243 &   &   &   &   \\\hline
\multirow[c]{2}{*}{pairwise} & P & \cellcolor{yellow}0.9681 & \cellcolor{yellow}0.1016 & \cellcolor{yellow}0.5506 & \cellcolor{yellow}0.3223 & \cellcolor{yellow}0.7946 & \cellcolor{yellow}0.5155 & \cellcolor{yellow}0.3938 & \cellcolor{yellow}0.2417 & \cellcolor{yellow}0.9266 & \cellcolor{yellow}0.0746 & \cellcolor{yellow}0.4108 & \cellcolor{yellow}0.9789 \\
 & N & \cellcolor{yellow}0.7013 & \cellcolor{yellow}0.0964 & \cellcolor{yellow}0.8638 & \cellcolor{yellow}0.0571 & \cellcolor{yellow}0.6871 & \cellcolor{yellow}0.3874 & \cellcolor{yellow}0.8669 & \cellcolor{yellow}0.3273 & \cellcolor{yellow}0.842 & \cellcolor{green}0.0131 & \cellcolor{yellow}0.5402 & \cellcolor{yellow}0.6781 \\\hline
\multirow[c]{2}{*}{sca} & P & \cellcolor{yellow}0.3545 & \cellcolor{yellow}0.4493 & \cellcolor{yellow}0.2169 & \cellcolor{yellow}0.6482 & \cellcolor{yellow}0.2225 & \cellcolor{yellow}0.4811 & \cellcolor{red}0.0006 & \cellcolor{red}0.0011 & \cellcolor{yellow}0.3923 & \cellcolor{yellow}0.3062 & \cellcolor{red}0.001 & \cellcolor{yellow}0.101 \\
 & N & \cellcolor{yellow}0.3656 & \cellcolor{yellow}0.5908 & \cellcolor{red}0.0017 & \cellcolor{yellow}0.1153 & \cellcolor{yellow}0.315 & \cellcolor{yellow}0.0654 & \cellcolor{red}0.0003 & \cellcolor{red}0.0022 & \cellcolor{yellow}0.8758 & \cellcolor{yellow}0.6311 & \cellcolor{red}0.0073 & \cellcolor{yellow}0.3553 \\\hline
  & & \multicolumn{4}{c|}{pairwise} & \multicolumn{4}{c|}{sca} \\\hhline{----------}
\multirow[c]{2}{*}{circular} & P & \cellcolor{yellow}0.9681 & \cellcolor{yellow}0.1016 & \cellcolor{yellow}0.5506 & \cellcolor{yellow}0.3223 & \cellcolor{yellow}0.3545 & \cellcolor{yellow}0.4493 & \cellcolor{yellow}0.2169 & \cellcolor{yellow}0.6482\\
 & N & \cellcolor{yellow}0.7013 & \cellcolor{yellow}0.0964 & \cellcolor{yellow}0.8638 & \cellcolor{yellow}0.0571 & \cellcolor{yellow}0.3656 & \cellcolor{yellow}0.5908 & \cellcolor{green}0.0017 & \cellcolor{yellow}0.1153 \\\hhline{----------}
\multirow[c]{2}{*}{full} & P & \cellcolor{yellow}0.7946 & \cellcolor{yellow}0.5155 & \cellcolor{yellow}0.3938 & \cellcolor{yellow}0.2417 & \cellcolor{yellow}0.2225 & \cellcolor{yellow}0.4811 & \cellcolor{green}0.0006 & \cellcolor{green}0.0011 \\
 & N & \cellcolor{yellow}0.6871 & \cellcolor{yellow}0.3874 & \cellcolor{yellow}0.8669 & \cellcolor{yellow}0.3273 & \cellcolor{yellow}0.315 & \cellcolor{yellow}0.0654 & \cellcolor{green}0.0003 & \cellcolor{green}0.0022\\\hhline{----------}
\multirow[c]{2}{*}{linear} & P & \cellcolor{yellow}0.9266 & \cellcolor{yellow}0.0746 & \cellcolor{yellow}0.4108 & \cellcolor{yellow}0.9789 & \cellcolor{yellow}0.3923 & \cellcolor{yellow}0.3062 & \cellcolor{green}0.001 & \cellcolor{yellow}0.101 \\
 & N & \cellcolor{yellow}0.842 & \cellcolor{red}0.0131 & \cellcolor{yellow}0.5402 & \cellcolor{yellow}0.6781 & \cellcolor{yellow}0.8758 & \cellcolor{yellow}0.6311 & \cellcolor{green}0.0073 & \cellcolor{yellow}0.3553 \\\hhline{----------}
\multirow[c]{2}{*}{pairwise} & P &   &   &   &   & \cellcolor{yellow}0.5738 & \cellcolor{yellow}0.259 & \cellcolor{yellow}0.1396 & \cellcolor{yellow}0.1724 \\
 & N &   &   &   &   & \cellcolor{yellow}0.7987 & \cellcolor{green}0.0226 & \cellcolor{green}0.0137 & \cellcolor{yellow}0.356 \\\hhline{----------}
\multirow[t]{2}{*}{sca} & P & \cellcolor{yellow}0.5738 & \cellcolor{yellow}0.259 & \cellcolor{yellow}0.1396 & \cellcolor{yellow}0.1724 &   &   & &  \\
 & N & \cellcolor{yellow}0.7987 & \cellcolor{red}0.0226 & \cellcolor{red}0.0137 & \cellcolor{yellow}0.356 &   &   &   & \\\hhline{----------}
\end{tabular}
    \caption{Ansatz Entanglement: Significance Tests}
    \label{tab:ans_ent_sign1}
\end{table*}

\end{document}